\documentclass{ws-igtr}
\usepackage{float}
\usepackage[colorlinks=true,citecolor=blue,linkcolor=blue]{hyperref}

\begin{document}


\catchline{}{}{}{}{}

\title{IS A GOOD OFFENSIVE ALWAYS THE BEST DEFENSE?}

\author{J. QUETZALCOATL TOLEDO-MARIN}
\address{Depto. de F\'isica Qu\'imica, Instituto de F\'isica, Universidad Nacional Aut\'onoma de M\'exico,\\
Apartado Postal 20-364, 
Mexico DF 01000, Mexico\\
j.toledo@fisica.unam.mx}

\author{ROGELIO DIAZ-MENDEZ}
\address{icFRC, IPCMS (UMR 7504) and ISIS (UMR 7006), Universit\'e de Strasbourg and CNRS,\\
CP 67000 Strasbourg, France\\
diazmendez@unistra.fr}

\author{MARCELO DEL CASTILLO MUSSOT}
\address{Depto. de Estado S\'olido, Instituto de F\'isica, Universidad Nacional Aut\'onoma de M\'exico,\\
Apartado Postal 20-364, 
Mexico DF 01000, Mexico\\
mussot@fisica.unam.mx}

\maketitle


\begin{abstract}
A checkers-like model game with a simplified set of rules is studied through extensive simulations of agents with different
expertise and strategies.
The introduction of complementary strategies, in a quite general way, provides a tool to mimic the basic ingredients of a
wide scope of real games.
We find that only for the player having the higher offensive expertise (the \emph{dominant player}), maximizing the
offensive always increases the probability to win.
For the non-dominant player, interestingly, a complete minimization of the offensive becomes the best way to win in many
situations, depending on the relative values of the defense expertise.
Further simulations on the interplay of defense expertise were done separately, in the context of a fully-offensive
scenario, offering a starting point for analytical treatments.
In particular, we established that in this scenario the
total number of moves is defined only by the player with the lower
defensive expertise. 
We believe that these results stand for a first step towards a new way to improve decisions-making in a large number of
zero-sum real games.
\end{abstract}

\ccode{PACS Nos.: 01.80.+b, 07.05.Tp, 02.50.Le}

\section{Introduction}
\label{int}

Expertise and strategy are the keystones for many sports. 
Expertise has to do with the degree of ability that a single player or team has, in order to perform an offensive or
defensive move. 
Strategy is a rule that associates a player's move with the information available to him at the time when he decides which
move to choose (\cite{haurie00}).
Now consider a real zero-sum game such as baseball, football or volleyball. 
These are strategic games (\cite{lam09}), since the teams (for simplicity called players from here on) choose their average
performance, i.e. more offensive or more defensive actions, once and for all and simultaneously.
This happens when  a particular line-up (and therefore a particular strategy) is chosen at the beginning of every match,
thus defining one of the many possible particular balances between defensive and offensive performance along the match.
A common belief on strategy, particularly in the context of simple games, is expressed with the adage ``the
best defense is a good offensive''.
In this work we test this belief through the construction of a properly designed checker-like model game. 

Checkers is a table game that occupies a very fundamental place in game theory for an important reason: it is the most
complex game ever solved (\cite{schaeffer07}).
It has been proved by extensive numerical calculations that, for two checkers players making no wrong moves, the game
always ends in a draw.
Indeed, checkers have been in the center of an intense research concerning machine learning and artificial intelligence
since the beginning of 1950s (\cite{schaeffer97}). 
The computational proof that checkers is a draw (\cite{cho07}), highlights the promising use of specialized algorithms in
statistical calculations of checker-like model games, and reinforces its conceptual similarity to many apparently dissimilar
games at first glance, such as team sports like baseball, football, etc.
Therefore the use of checkers as a model game, with the smaller quantity of ingredients needed to develop a general
phenomenology, is a well based approach.
The study of simplified game models allows to better generalize statistical results to near-connected games, as well as to
physical and economic processes (\cite{haurie00}).
With this aim, we have built a smart program that extensively simulates a variant of the checkers game, looking for easily
generalizable regularities that are impossible to asses with more specific game designs.

Extensive statistical explorations have proved to be a valuable and consistent way to establish game properties, and validate
usual assumptions in game theory (\cite{aucamp86,rodriguez06,marcelo}).
We perform such exploration in our model game implementing a complementary strategy, that is likely to mimic real games in a
general way.
In the complementary strategy scenario, a single value is defined to indicate the balance between the offensive and defensive
performances, in such a way that players cannot maximize both at the same time.
Thus, we ran simulations looking for optimal values in the whole range of possible strategies, considering arbitrary
different expertise values (both defensive and offensive) between opponents.
As a result, we find that the common belief on maximizing offensive is only true for what we call the {\it dominant
player}, i.e. that of the highest offensive expertise. 
For the non-dominant player, maximizing offensive can have in some cases the opposite effect depending on the defensive expertise values. 
This proof is of remarkable interest, not only as a useful prospect for guiding decision-making, but also because it
highlights the importance of a reliable assessment of the expertise prior to the selection of the strategy.

The fact that the best strategy is non-trivial for the non-dominant player, raises the role of the defensive expertise.
We thus addressed the isolated effects of defensive expertise in a purely offensive scenario, as a starting point for a
further analytical treatment of the simple model game.
In this extreme situation, several quantities were found to be instrumental for both time-dependent and time-independent
game observables.
In particular we find the quantitative relation of the defensive expertise value with the total time length of the match,
the growing of advantages and the distribution of even sequences. 

The rest of the paper is organized as follows: 
In the next Section~\ref{model} we introduce the model, its particular features, and the definition of the offensive and
defensive expertise. 
In Section~\ref{cs} we present and discuss the main results of the work, concerned with the complementary strategy implementation.
In Section~\ref{fos} we focus on the effects of defensive expertise through the study of the fully-offensive scenario as an
extreme case.
Section~\ref{c} is devoted to the general conclusions of the work.

\subsection{Model}
\label{model}

The simulations are done on a standard checker scheme: twelve pieces for each player in a conventional board of
$8\times8$ sites, the half of which may be occupied.
In the same way, the capturing procedure and the criterion to finish the game, were implemented as in standard
checkers~(\cite{bc2010}). 
However, in contrast to the standard rule according to which moves are only valid in the two forward diagonal directions,
here we allow moves in the four diagonal directions (provided the nearest site is empty).
Two further simplifications were considered: only one capture is allowed for a single move, and the king crowning is
forbidden.
The match starts randomly with one of the two players, and the turn to move alternates in the following, until one of them
runs out of pieces and his opponent becomes the winner.  

\begin{figure}[!ht]
\begin{center}
\includegraphics[width=0.45\textwidth ]{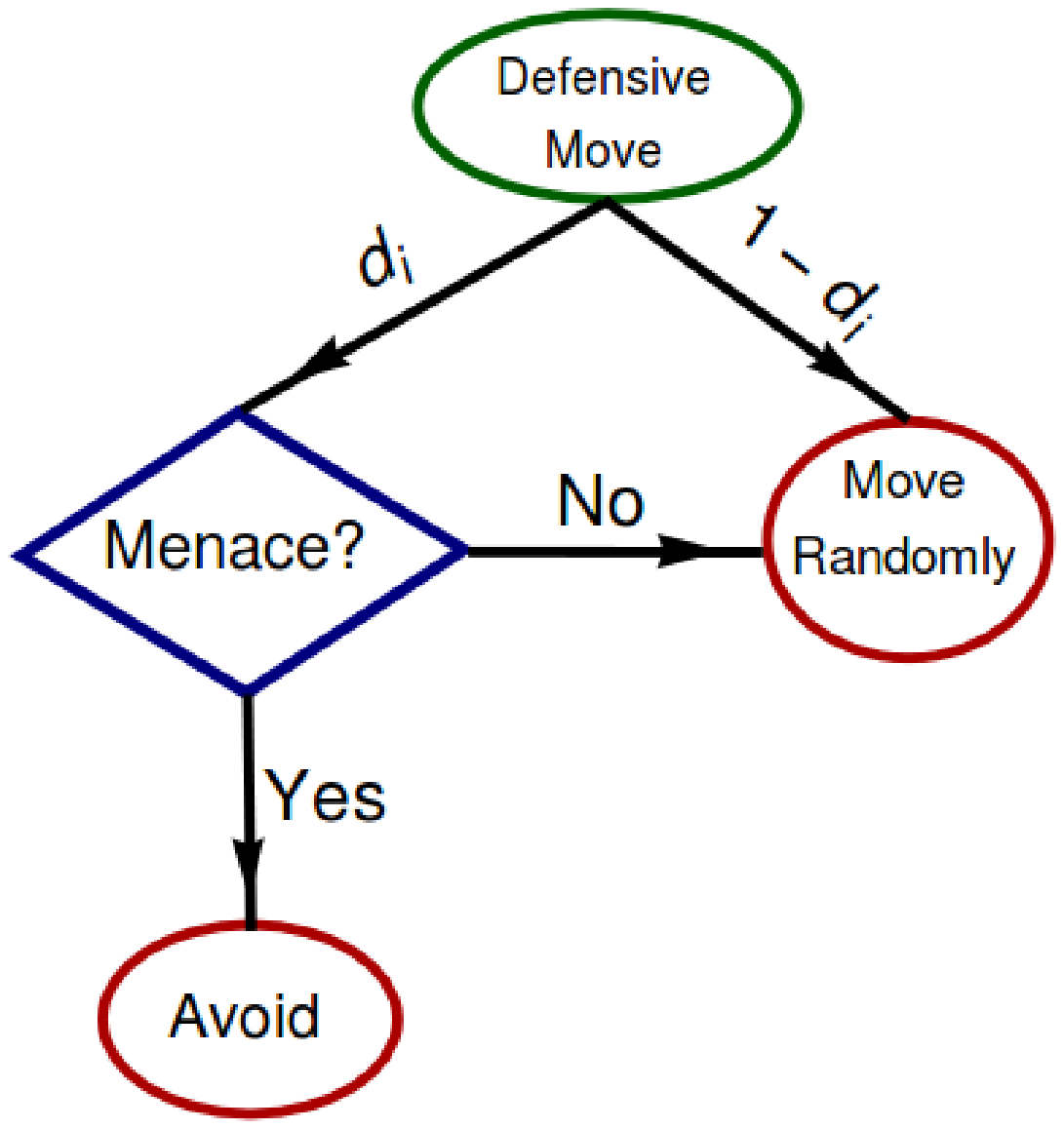}
\includegraphics[width=0.45\textwidth ]{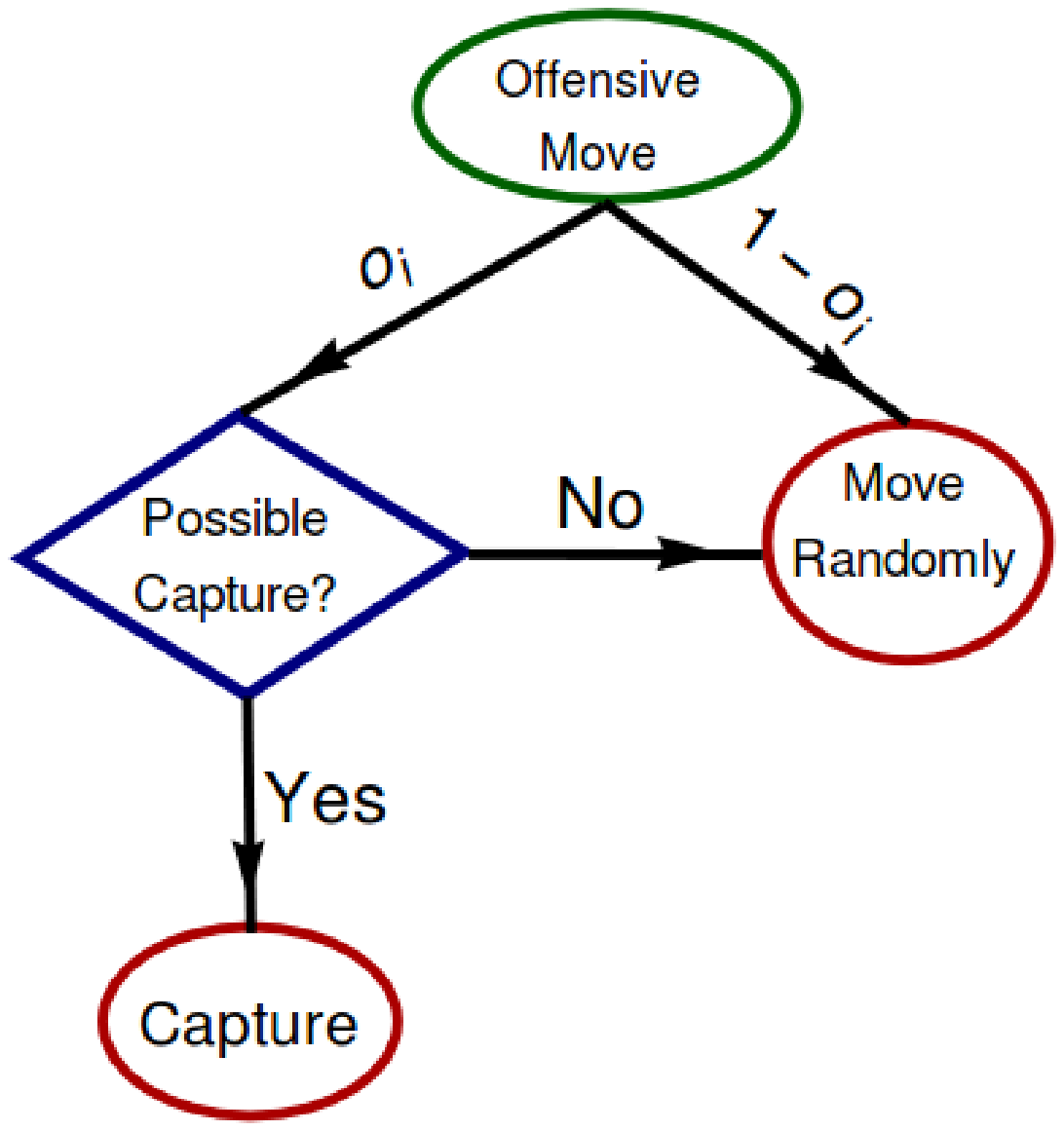}
\end{center}
\caption{Decision trees corresponding to the defensive (left) and offensive (right) moves.}
\label{fig:SimDiagram}
\end{figure}

Each player $i$, with $i\in\lbrace 1,2 \rbrace$, has two associated values representing his defensive expertise $d$ and
offensive expertise $o$.
We will refer to them as $d_i$ and $o_i$, and both are continuous variables ranging from $0$ to $1$. 
Defensive expertise $d_i$ stands for the capacity of player $i$ to avoid menaces, if any: when a defensive move is requested
for player $i$, he will look, with a probability $d_i$, for menaced pieces, and (in case there exist) move one of them to
avoid the menace.
Correspondingly, the offensive expertise $o_i$ stands for the capacity of player $i$ to perform captures. 
Once an offensive move is commanded to player $i$, he will check, with a probability $o_i$, if some opponent piece can be
captured, and if so he will make the move.

Figure~\ref{fig:SimDiagram} shows the decision tree corresponding to the defensive and offensive moves.
Note that if $d_i=0$, player $i$ will perform random moves whenever a defense move is requested; while if $d_i=1$, he will
then evaluate the menaces and avoid them whenever it is possible.
Any value in between $0<d_i<1$ means that the player will avoid menaces with probability $d_i$, or move randomly with
probability $1-d_i$.
Analogously, a player $i$ with $o_i=0$, asked to move offensively, will actually move randomly; while if $o_i=1$ the player
will always capture an opponent piece, provided there is a piece to capture. 
For $0<o_i<1$ this player will evaluate (and perform) captures with probability $o_i$ or move randomly otherwise.

The particular choice a player makes in every turn, that is, the choice of performing a defensive move, an offensive move, or
any other, is governed by its strategy. 
In this way, the input of its expertise and strategy completely defines how a player performs in a match. 


\section{Complementary strategy}
\label{cs}

The motivation behind the complementary strategy can be easily understood if we go back to our baseball game example. 
Prior to the beginning of the match, a line-up of nine individuals has to be chosen among the whole team, this
line-up, aside from minor changes, will be kept throughout the match. 
As we already pointed, this is equivalent to the choice of a particular strategy. 
In a good approximation, one can consider that the line-up with the best offensive performance is not likely to be the
same as the one with the best defensive performance.
Indeed, individuals conforming the team are generally specialized in one of the two performances.
In this way, a particular line-up maximizing the offensive performance will minimize the defensive performance, and vice
versa.

\begin{figure}[!ht]
\begin{center}
\includegraphics[width=0.7\textwidth ]{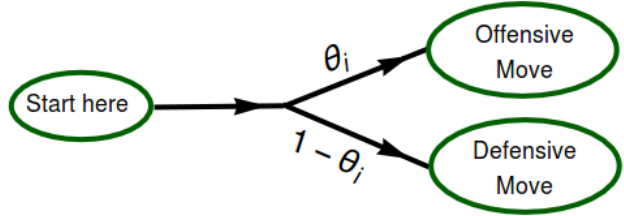}
\end{center}
\caption{Decision tree corresponding to the complementary strategy.}
\label{fce}
\end{figure}

For our model game, we define the complementary strategy as a number $0\leq\theta\leq1$ according to which in every
turn a player $i$ makes an offensive move with a probability $\theta_i$, or a  defensive move with a probability
$1-\theta_i$ (see fig. \ref{fce}).
At the beginning, each player is provided with a fixed value of $\theta_i$ that remains constant throughout the match.
We notice that, with this definition, for $\theta_i=1$ or $\theta_i=0$ the complementary strategy corresponds to
a pure strategy, while otherwise it becomes a mixed strategy.

When using the complementary strategy, the winning rate of the players strongly depends on the particular choice of
$\theta_1$ and $\theta_2$.
For a fixed set of expertise values $\lbrace o_1,d_1,o_2,d_2 \rbrace$, we fix the strategies $\theta_i$ and simulate $10^5$
matches with different random seeds.
We then change the strategy to a new value in steps of $\Delta\theta=0.05$, exploring all the range of $\theta$ between $0$
and $1$. 
The outcome resulting from the statistical treatment of the data is used to draw the winning matrices
corresponding to the given set of expertise values.

\begin{figure}[!ht]
\begin{center}
\includegraphics[width=0.45\textwidth ]{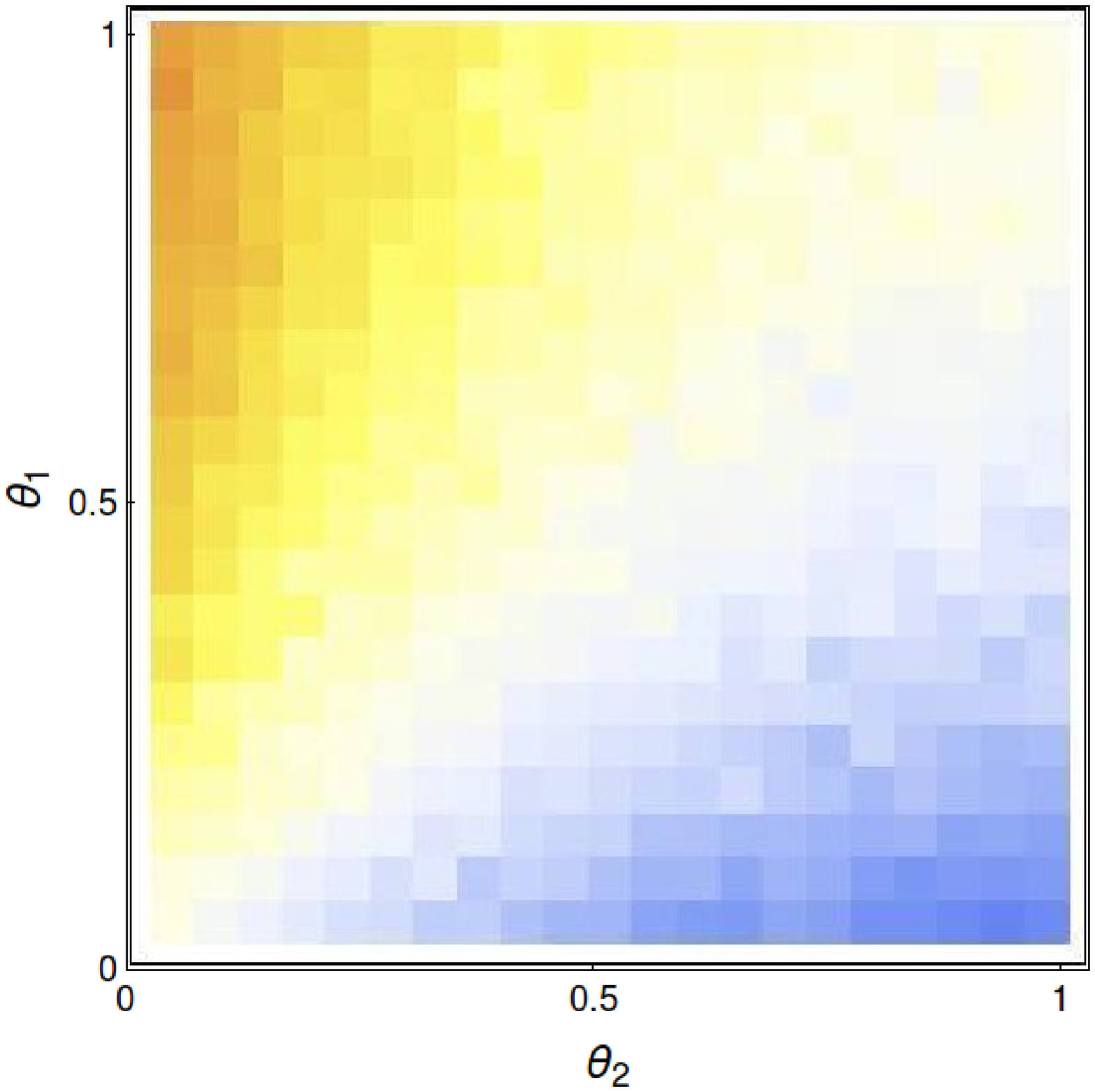}
\includegraphics[width=0.45\textwidth ]{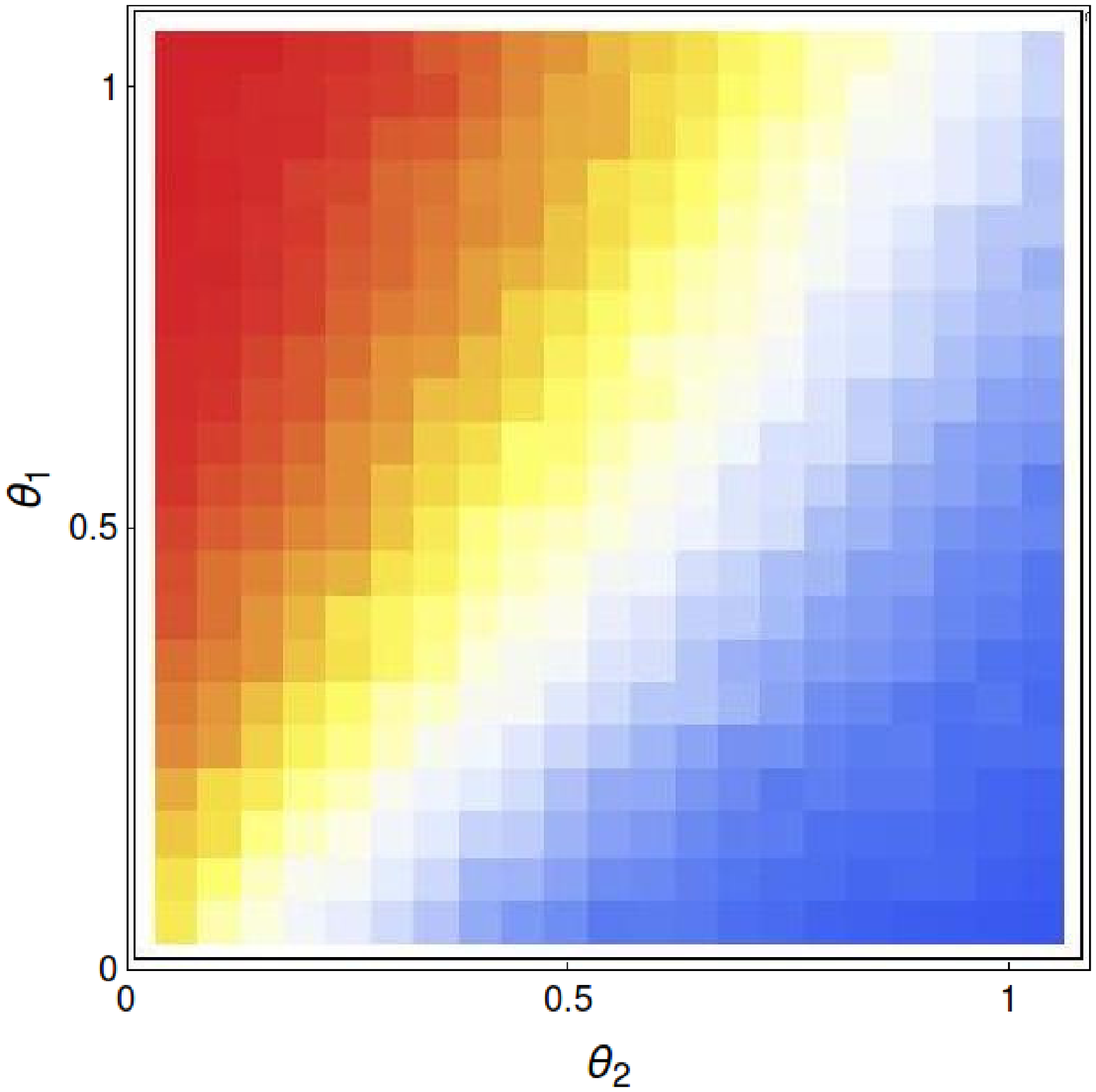}
\end{center}
\caption{Wining matrix for two different sets of expertise values.
Left:   $d_1=d_2=0.75$ and $o_1=o_2=0.75$.
Right:  $d_1=1$, $o_1=0.25$ and $d_2=0.25$, $o_2=0.75$.
In each panel, strategy $\theta_1$ grows from bottom to top and $\theta_2$ grows from left to right; red (blue) intensity
indicates that player 1 (2) has the highest winning rate.
}
\label{m1}
\end{figure}

\begin{figure}[!ht]
\begin{center}
\includegraphics[width=0.45\textwidth]{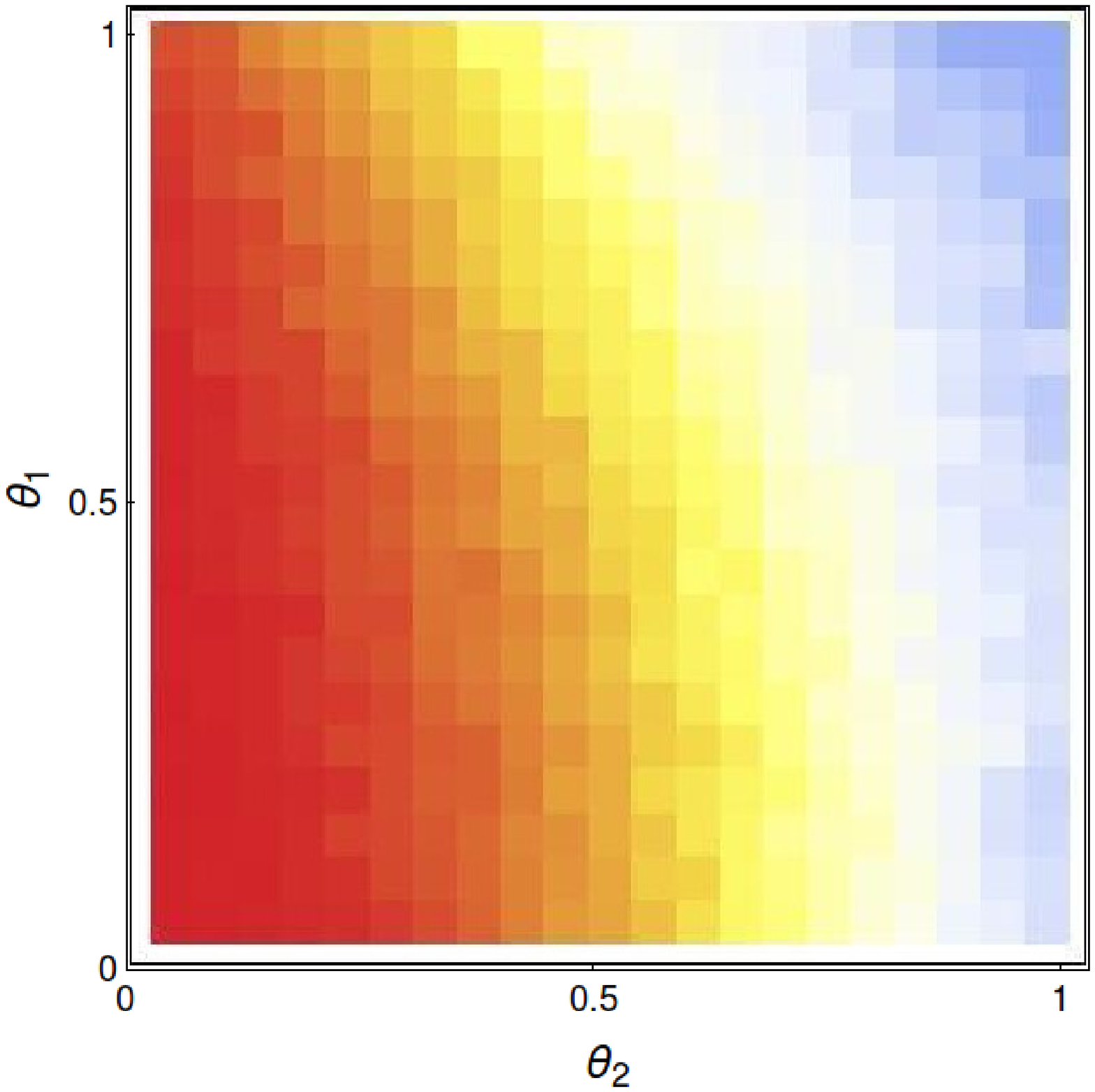}
\includegraphics[width=0.45\textwidth]{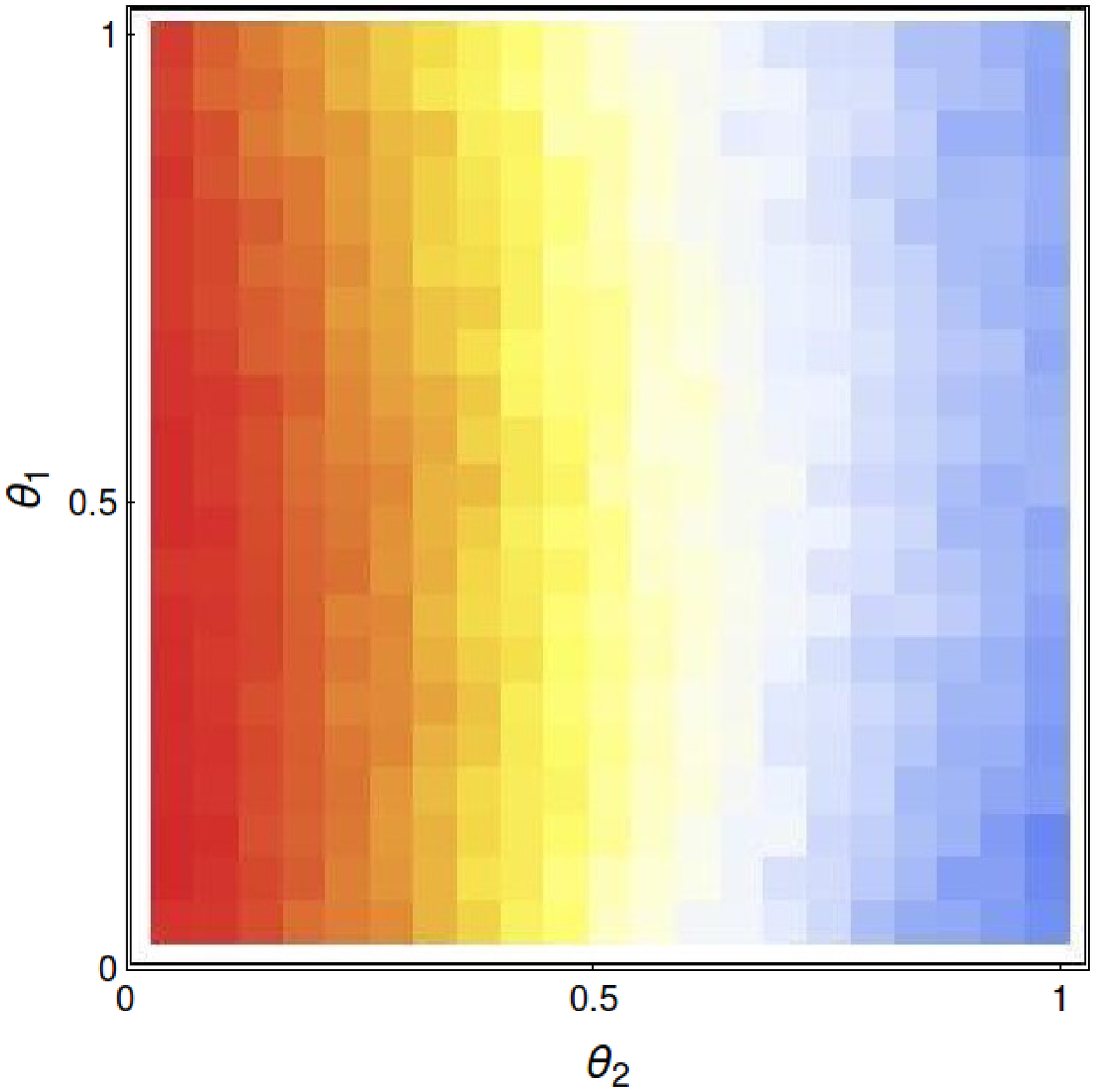}
\end{center}
\caption{Wining matrix for two different sets of expertise values.
Left:  $d_1=1$, $o_1=0.5$ and $d_2=0.25$, $o_2=0.75$.
Right: $d_1=0.75$, $o_1=0.5$ and $d_2=0.25$, $o_2=0.75$.
In each panel, strategy $\theta_1$ grows from bottom to top and $\theta_2$ grows from left to right; red (blue) intensity
indicates that player 1 (2) has the highest winning rate.
}
\label{m2}
\end{figure}

In a winning matrix, every row (column) represents a specific value of the strategy $\theta_1$ ($\theta_2$), which
corresponds to player 1 (2).
A given element of such matrix is calculated as the difference between the number of matches won by player 1 and those
won by  player 2.    
In Fig.~\ref{m1} and  Fig.~\ref{m2} the value of the matrix elements has been normalized and represented in the form of a
temperature map, ranging from -1, which is indicated by dark blue, to 1 corresponding to dark red.
In this way, red (blue) intensity indicates that a match under these conditions is more probably won by player 1 (2). 
The white color means that the winning rate is the same for both players.

As can be seen from figure \ref{m1}~(left), the winning rate for players with the same expertise values
increases as the strategy becomes more offensive.
This matrix is representative of those games in which players have the same expertise values. 
Under this condition, the common belief concerning the maximization of the offensive is indeed the best strategy,
i.e., $\theta_i=1$.
Furthermore, for the player with the highest offensive expertise, all further outcomes consistently showed that this
offensive maximization leaded to the highest winning rate.  
In other words, if $o_i\geq o_j$ then the best strategy for player $i$ is always $\theta_i=1$. 
In the following, the player with the highest offensive expertise is called the {\it dominant player}.

In the case presented in figure~\ref{m1}~(right), we can appreciate that the offensive maximization strategy is the best choice not only for the
dominant player (player 2), but also for the non-dominant player (player 1).
Notice, from the last column of the winning matrix in figure~\ref{m1}~(right), that even when the dominant player can statistically ensure the
winning by
choosing $\theta_2=1$, the non-dominant player 1 can still expect to improve his odds by setting $\theta_1=1$.
However, as we show below, for the non-dominant player, this offensive maximization is not an
universal recipe of improvement.

The exploration of a wide range of non-equal expertise values, revealed a further richer scenario concerning the best
strategy for the non-dominant player.
In figure~\ref{m2}~(left) a representative case is shown, where  player 2, with winnings in blue, is the dominant
player. 
Consequently, it is always possible to select a strategy $\theta_2=1$ for which the winning rate favors player 2, no
matter which strategy player 1 chooses. However, in this case, the non-dominant player 
has better odds at winning if he chooses a strategy which maximizes his defensive performance.
This type of situation is non-trivial, and its implication for decision-making is remarkable.
In this situation, for instance, if a non-dominant baseball team (player 1) decides to maximize its defensive performance,
the dominant team (player 2) will be forced to fully enhance its offensive performance in order to slightly tip the balance
in his favor. 
Actually, player 1 has the freedom of setting a strategy $\theta_1\leq0.5$ to reach the same goal, since the matrix
values are virtually the same in this range. 
Figure~\ref{m2}~(right) shows the example of a case in which the match behaves roughly in the same way for all possible strategy
values of the non-dominant player.

In order to recreate these situations presented in figure~\ref{m2}, a necessary condition is that the defensive expertise $d$ of the
non-dominant player has to be larger than that of the dominant player.
Thus, even though the dominant player (defined through his offensive expertise) can always find a strategy that favors
his winning rate, the shape of the matrices is largely influenced by the role of defensive expertise values.
While the winning matrices in figure~\ref{m1}~(right) and figure~\ref{m2}~(left and right) correspond to the same dominant player
$d=0.25; o=0.75$, the best strategy is different for each of the three different opponents.
Moreover, in the three cases the dominant player has to face very different challenges:
in the situation presented in Fig.~\ref{m1}~(right), he has to adopt a strategy value larger than that of the opponent; 
in the situation presented in Fig.~\ref{m2}~(right), he has to adopt a strategy value larger than about $0.6$; 
and in the situation presented in Fig.~\ref{m2}~(left) he has to adopt a full offensive maximization.

In terms of real game scenarios, these translate into very tough situations, since the assumption that all values of the
strategy are equally accessible is almost never the case in real sports.
As one of many examples, the dynamics of a baseball tourney makes it impossible to keep the same
line-up from one game to the next, thus dramatically restricting the available values of $\theta$. 
In this common situations the knowledge of the winning matrices may be crucial in planning a long term behavior.

\section{Role of defensive expertise: fully-offensive strategy}
\label{fos}

\begin{figure}[!ht]
\begin{center}
\includegraphics[width=0.9\textwidth ]{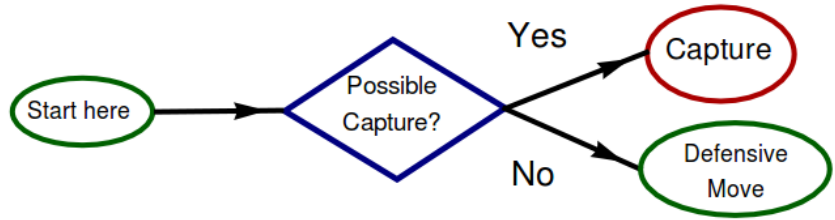}
\end{center}
\caption{Decision tree corresponding to the fully-offensive strategy.}
\label{fig:SimDiagram2}
\end{figure}

In order to better understand the effect of defensive expertise, in this section we implement the fully-offensive strategy,
as an artificial extreme case.  
In this scenario, the offensive expertise values are fixed at the maximum value, i.e., $o_1=o_2=1$, while the range of defensive
expertise $d_1\geq d_2$ is carefully explored.
The fully-offensive strategy, shown in Fig.~\ref{fig:SimDiagram2}, consists in a fixed algorithm for every move and is
exactly the same in both players: on each move, the players will always capture a piece provided there is one to be captured,
or make a defensive move otherwise.
This capture at the beginning of the tree in Fig.~\ref{fig:SimDiagram2} allows to equally neglect the influence of the
offensive performance. 
Although this strategy is an artificial extreme case, and does not correspond to any real game format, we show it is very
useful to study the isolated effects of the defense expertise.

We carried out simulations for opponents with defensive expertise $d~\in~\lbrace 0,0.5,0.75,1\rbrace$. 
For each combination we ran up to $10^6$ matches in order to have a suitable statistics. 
In the fully-offensive frame, matches with $d_1=d_2=1$ are quite difficult to end, since the players move avoiding menaces
whenever they appear, thus producing a dramatic increase in the length of the match as the board becomes less occupied.
We do not account for this particular combination of expertise values in our study.

The analysis in fully-offensive strategy was divided in two sections devoted to the study of time-dependent and
time-independent behaviors.
Time $t$ is represented here by the sequence of moves, in such a way that a move of player 1 plus a move of player 2 are two
units of $t$. 
In the time-dependent study we focused in the total time of the matches $\tau$, and the material advantage
$v(t)$ (\cite{marcelo}), defined as the difference between the number of pieces of player 1 and player 2.
The time-independent study focused in the length of even sequences $L$, also measured in number of moves.

\subsection{Time-dependent properties}
\label{tdp}

\begin{figure}[!htb]
\includegraphics[height=0.49\textwidth, angle=-90]{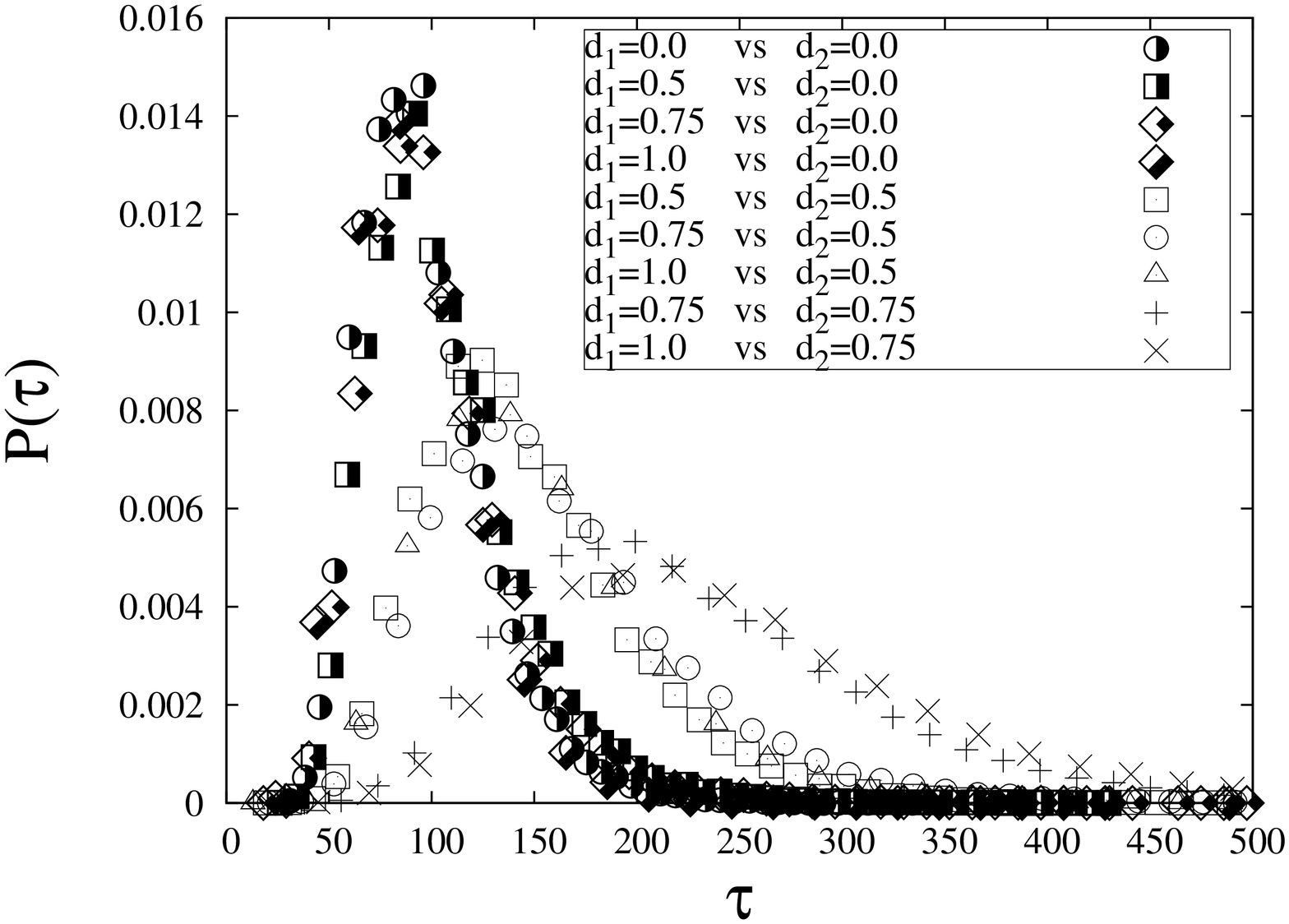}
\includegraphics[height=0.49\textwidth, angle=-90]{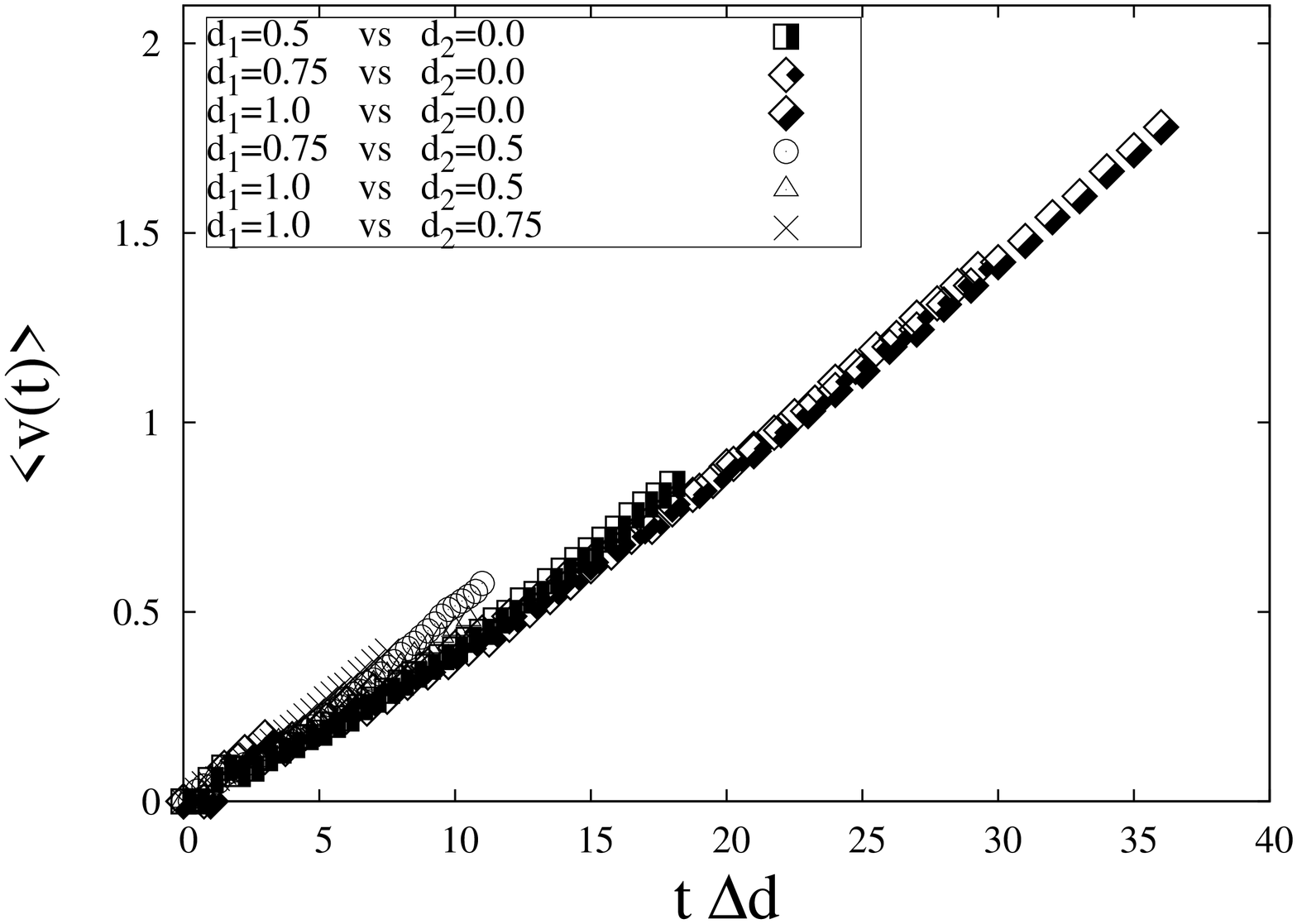}
\caption{Left: Histograms of the total time $\tau$ for several combinations of expertise $d_1\geq d_2$.
Partially-filled symbols represent matches where the lower expertise is $d_2=0$, for void  symbols $d_2=0.5$ and for open
symbols $d_2=0.75$.
Right: Averaged advantages for the initial stages of matches with several combinations of expertise values.
Collapse of the averaged advantage with $t \Delta d$.
}
\label{taus-ex}
\end{figure}

In the left panel of figure \ref{taus-ex} we have plotted histograms for the total time of the matches $\tau$, for
several combinations of expertise values. 
Interestingly, when the lowest expertise ($d_2$) is the same, all the histograms collapse into a single, well-defined curve.
In other words, the distribution of total time $\tau$ is independent of the value of the highest expertise.
Thus, the total time of a single game can be completely described in terms of the worst player, and his defensive
expertise value $d_2$ is the only relevant parameter influencing the behavior of games in terms of length.

The mean total time of the matches increases by increasing $d_2$, as can be observed from the shift to the right in
the histograms of figure \ref{taus-ex} (left).
This increase does not happens in a trivial way, 
but accompanied of
a marked increase in the dispersion and keeping relatively constant the value of
the lower limit of the histograms. 
That is to say, the better the worst player is, the more difficult it is to predict the length of a single match.
On the other hand, the lower limit in the histograms of total time, is associated to the fact that there is a minimal number
of moves needed to capture all the pieces.
This number is about $\tau_\mathrm{min}\approx30$, corresponding to $2.5$ moves per piece in average.

The calculation of advantages $v(t)$ was performed by the simple difference of pieces of player 1 ($+1$) and pieces of player
2 ($-1$) at every time (turn) of the match. 
By definition, when $v>0$ ($v<0$) player 1 (2) has more pieces than player 2 (1).
With the aim of statistically studying the way in which  advantages departs from zero, the average $\langle v(t)\rangle$ over
the realizations is done only for initial times $t\leq\tau_\mathrm{min}^{d_1,d_2}$, so including the total
number of matches at every time $t$. 
The behavior of $\langle v(t)\rangle$ can be seen in the right panel of figure \ref{taus-ex} for several values of
combinations $d_1\neq d_2$.
In figure \ref{taus-ex} (right) a collapse is obtained by plotting the averaged advantage as a function of $t \Delta d$, 
where $\Delta d=d_1-d_2$. 
Thus, the mean advantages behaves linearly with time, and its growing speed is determined just by the difference in
defensive expertise between the two  opponents.

\subsection{Time-independent properties}
\label{tip}

The length of even sequences $L$ is the time span of a single value in the advantage $v$, i.e. a sequence of moves in which
equilibrium establishes. 
It is computed  without taking into account the quick alteration of two consecutive (and opposite) captures.
In that
case advantage suddenly returns to its previous value and it may be interpreted as an exchange, in analogy to
well-known plays existing in many games.

\begin{figure}[!ht]
\includegraphics[height=0.49\textwidth, angle=-90]{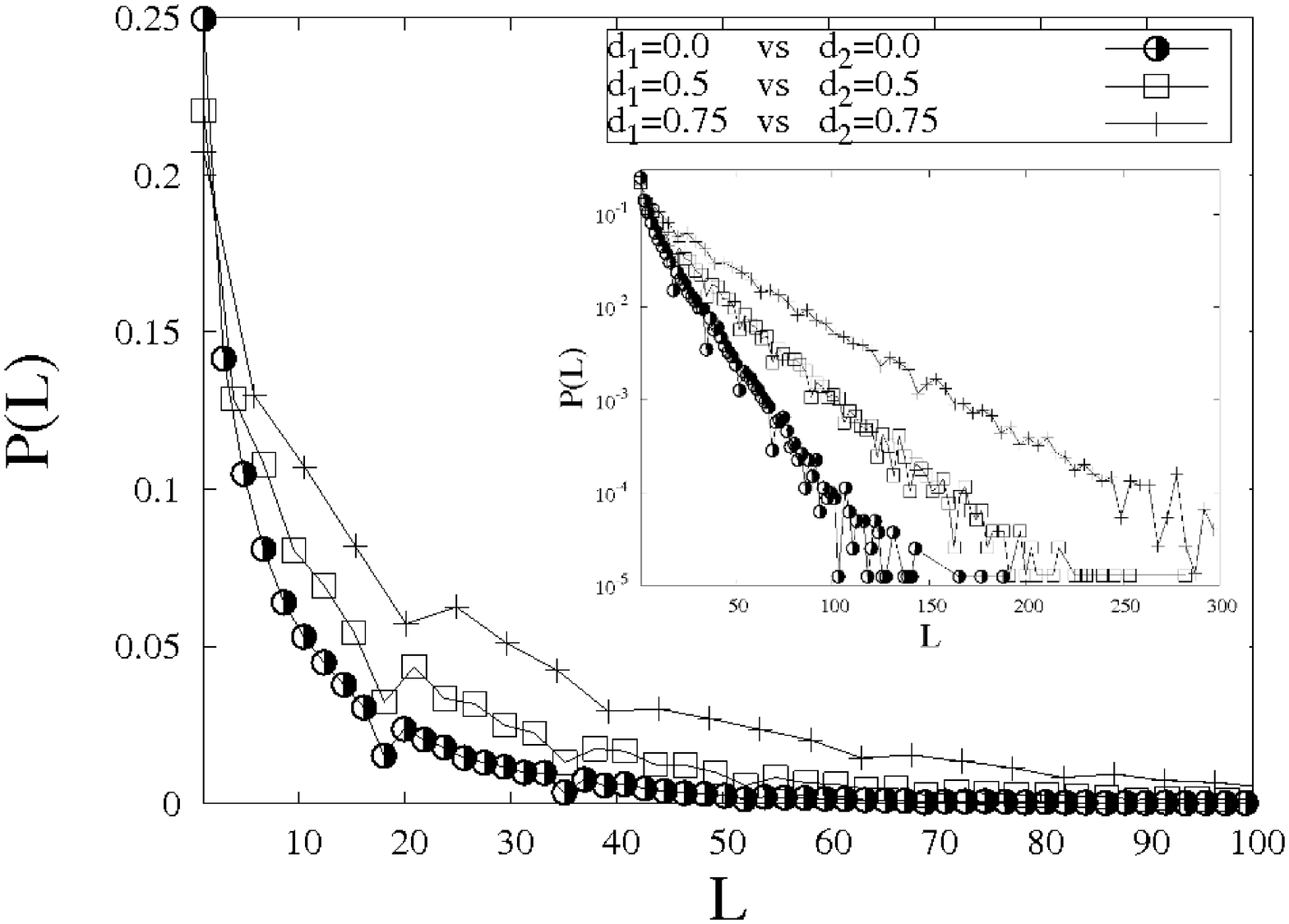}
\includegraphics[height=0.49\textwidth, angle=-90]{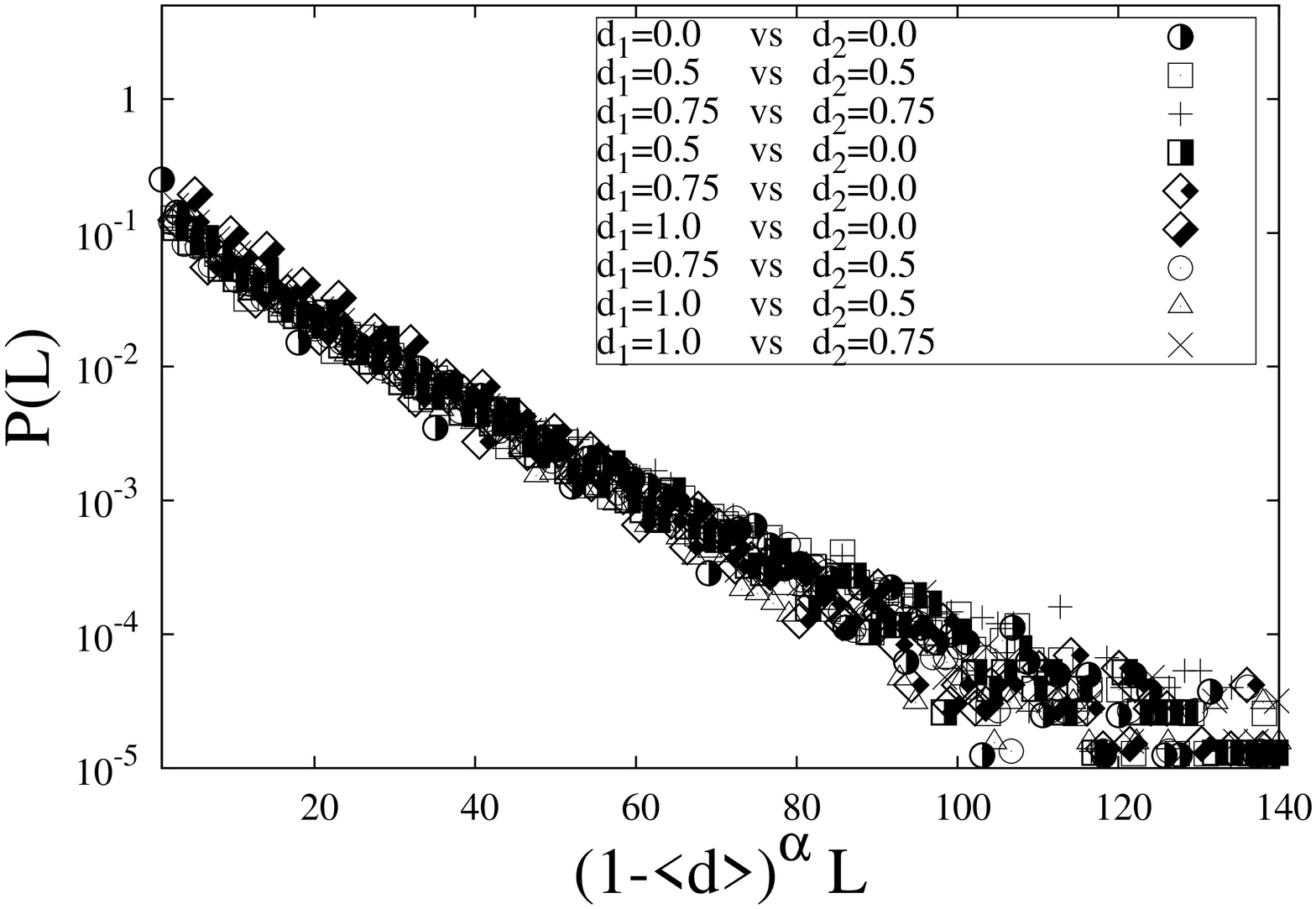}
\caption{Left: Histograms of the length of even sequences in matches between opponents with the same expertise. 
The inset is a logarithmic scaling of the y-axis in order to remark exponential behaviors.
Right: Collapse of the distributions of the length of even sequences with the functionality (\ref{lbd}) for all values of
$d_1$ and $d_2$. The graph corresponds to the best collapse with $\alpha=0.65$.}
\label{semp-sempc}
\end{figure}

In figure \ref{semp-sempc} (left) one can see the distribution of $L$ for matches between opponents of equal expertise.
While the more probable sequence is around $L=1$, it is evident that the higher the expertise is, the higher is the
probability of finding larger lengths in the game. 
Clearly, one expects that good players sustain larger even sequences than bad players.
This rather intuitive behavior appears in a wide variety of real games.

The exponential decay of the length distributions is another feature that can be seen from the inset of figure
\ref{semp-sempc} (left). 
This behavior is ruled by an expertise-dependent exponential factor 
$P(L)\sim e^{- \lambda(d_1,d_2) L}$
that does not depends on the expertise difference $\Delta d$. 
One natural assumption is to consider $\lambda$ as a function of the mean expertise $\langle d\rangle=(d_1+d_2)/2$ through
the form
\begin{equation}
\lambda(d_1,d_2)\sim (1-\langle d\rangle)^\alpha \; .
\label{lbd}
\end{equation}
In figure \ref{semp-sempc} (right) the data of length distributions is collapsed with this functionality. 
The value $\alpha=0.65$ produced a good collapse over several decades for all the values of $d_1$ and $d_2$ tested in this
work.
In this way, the length of even sequences can be explained in terms of the mean expertise of the opponents.
The latter shows the existence of another quantity through which defensive expertise mediate the
behavior in the fully-offensive approach.

\section{Conclusions}
\label{c}

We have presented an extensive numerical study on a simple model game with zero-sum. 
This type of game covers a wide scope of real sports for which a systematic study of the interplay between expertise and
strategy has not been definitely addressed.
With this motivation we implemented a complementary strategy that maximizes the offensive performance by
minimizing the defense performance, and vice versa, in a continuous way.
This approach emulates real situations as, for example, the choice of a particular line-up from a baseball team.
Far from simplicity, our model demonstrate that there is a rich scenario concerning winning expectations, that can be
visualized through the winning matrices.
The winning matrix is constructed by statistical simulations and allows the identification of the best strategies in a
number of different scenarios.

Our results suggest that the common belief that ``the best defense is a good offensive'' is only true for the dominant
player.
For the non-dominant player the best strategy to adopt is strongly dependent on the four expertise values (defensive and
offensive of the two players) and usually is that of minimizing offensive. 
For decision makers this is a valuable knowledge since the evaluation of expertise in real games can be done prior to the
selection of a strategy. 
This result makes a direct connection to the subject of decision-making based on reputation, which is rarely considered in
game analysis (\cite{lam09,mui02}).

Finally, a fully offensive strategy was also studied with the aim of quantitatively addressing the influence of defensive expertise in
an isolated context.
Statistical simulations of games between opponents of different expertise values showed that the total length of the matches
was determined only by the value of the expertise of the worst defensive player. 
The growth of advantages was proportional to the expertise difference between the opponents, and the distribution of even
sequences was determined only by the average expertise through an exponential law.
We consider these last results as a first step for a further analytical systematization.

\section{Acknowledgments}
\label{a}

M.C.M. thanks partial financial support from DGAPA-UNAM through grant IN105814. The authors would like to thank the referees
for their insightful comments and advices.

\end{document}